\pdfoutput=1

\documentclass[11pt]{article}

\usepackage[]{acl}

\usepackage{times}
\usepackage{latexsym}

\usepackage[T1]{fontenc}

\usepackage[utf8]{inputenc}

\usepackage{microtype}
\usepackage[tbtags]{amsmath}
\usepackage{graphicx}
\usepackage[linesnumbered,ruled]{algorithm2e}  
\title{Look Backward and Forward: Self-Knowledge Distillation with Bidirectional Decoder for Neural Machine Translation}

 \author{Xuanwei Zhang \and Libin Shen \and Disheng Pan \and Liang Wang \and Yanjun Miao \\ 
    $\left\{\rm {zhangxuanwei, shenlibin, pandisheng, liangwang, miaoyanjun}\right\}$@qiyi.com}
    
\begin{document}
\maketitle
\begin{abstract}
Neural Machine Translation(NMT) models are usually trained via unidirectional decoder which corresponds to optimizing one-step-ahead prediction. However, this kind of unidirectional decoding framework may incline to focus on local structure rather than global coherence. To alleviate this problem, we propose a novel method, \textbf{S}elf-Knowledge Distillation with \textbf{B}idirectional \textbf{D}ecoder for \textbf{N}eural \textbf{M}achine \textbf{T}ranslation(\textbf{SBD-NMT}). We deploy a backward decoder which can act as an effective regularization method to the forward decoder. By leveraging the backward decoder's information about the longer-term future, distilling knowledge learned in the backward decoder can encourage auto-regressive NMT models to plan ahead. Experiments show that our method is significantly better than the strong Transformer baselines on multiple machine translation data sets.
\end{abstract}


\section{Introduction}

Neural Machine Translation (NMT) \citep{bahdanau2015neural,bahdanau2016actor} is one of the most important tasks in the field of NLP and has achieved rapid development with the progress of deep learning research. NMT is essentially an encoder-decoder architecture\citep{koehn2003statistical}. In the conventional NMT model, firstly the source sequence is fed into the bidirectional encoder to learn contextual information, and the encoder takes in the input sequence (source language) and maps it to an intermediate hidden vector (higher dimensional space) which encodes all the information of the source.
This, in turn, is taken by the unidirectional decoder which generates an output sequence (target language) word by word. Meanwhile, In order to prevent over-ﬁtting and improve the generalization ability of deep models, there are many regularization techniques\citep{srivastava2014dropout,hinton2012improving,wan2013regularization,ioffe2015batch,ba2016layer,wu2018group,szegedy2016rethinking,hinton2015distilling,zhang2019your,krueger2016zoneout,krueger2015regularizing,merity2017regularizing}. Different from most of the previous regularization methods that act on the hidden states either by injecting noisee\citep{krueger2016zoneout} or by penalizing their norm \citep{krueger2015regularizing,merity2017regularizing}, SBD-NMT regularizing a neural machine translation network that encourages states of the forward decoder to predict cotemporal states
of the backward decoder.

 Due to the autoregressive structure, current NMT systems usually suffers from the so-called exposure bias problem\cite{zhang2019regularizing}: during inference, the previous word is not the golden word, but the word generated by the model. The situation mentioned above caused the complementary L2R(Left-to-Right) model tends to generate translation results with good prefixes and bad suffixes. Similar to using the L2R model, the R2L(Right-to-Left) model generates results with good suffixes and bad prefixes. Therefore, the translation errors during inference will be propagated to the task of translating the next word, resulting in unsatisfactory translation results. 
 With our proposed approach, reverse decoder’s looking into the future ability can act as an effective regularization method, capturing subtle long-term dependencies that ensure global coherence and in consequence boost model performance on machine translation.

To address the above problems, one line of research attempts to reduce the inconsistency between training and inference so as to imporve the robustness when giving incorrect previous predictions, such as
\citep{zheng2018modeling} introduced two additional recurrent layers to model translated past contents and untranslated future contents. \citep{he2016dual} applied a complicated reinforcement approach which use two agents to learn information from each other. During inference\citep{zhang2018asynchronous}, the process of predicting next word requires two-stage decoding, which is very time-consuming. \citep{zhang2019regularizing} fuses with L2R and R2L respectively, but the phase of training is complicated because of dynamic sampling. \citep{zhang2019future} distilled future knowledge from a backward neural language model trained only by target sequence data to provide full-range context information.

Instead of previous work, we propose a novel model regularization method for NMT training, which aims to improve the agreement between translations generated by L2R and R2L NMT decoders and learns better models integrating their advantages to generate translations with good prefixes and good suffixes. We integrate the optimization of R2L and L2R decoders into a joint training framework, in which they act as helper systems for each other. Meanwhile, our model does not need to change the conventional autoregressive decoding method during inference.


Therefore, our contributions in this work are as follow:
\begin{itemize}
 \item  We propose SBD-NMT, a simple yet effective regularization method by distilling the future-aware information contained in the backward decoder.

 \item Extensive experiments are conducted on multiple machine translation datasets. Experiments show that our proposed approach signiﬁcantly outperforms strong Transformer baselines on multiple translation datasets.
 

\end{itemize}






\section{Related Work}
Our research is built upon a sequence-to-sequence model \citep{sutskever2014sequence}, but it is also related to Bidirectional Decoding, Self-Knowledge Distillation. We will discuss these topics as follow.

\subsection{Bidirectional Decoding}
Many researches have been conducted to improve the translation quality through a bidirectional decoder. Backward language model or bidirectional decoding were first introduced into statistical machine translation (SMT) models in \citep{watanabe2002bidirectional,finch2009bidirectional,zhang2013beyond}, and achieved better performance.

Recently, \citep{liu2016agreement} and \citep{zhang2018asynchronous} migrated the method from SMT to NMT by modifying the inference strategy or the decoder architecture of NMT. \citep{liu2016agreement} proposed to generate ${N}$ best translation candidates from L2R and R2L NMT models, separately. Then used a joint model to rank the merged ${2N}$ candidates to find the best candidate. In 2018, \citep{zhang2018asynchronous} equip the conventional attentional encoder-decoder NMT framework with a backward decoder to explore bidirectional decoding. When predicting at each time step, the forward decoder simultaneously applies two attention models to consider the hidden state of the source and the backward target respectively. 

The work most related to ours is a Synchronous Bidirectional Neural Machine Translation(SB-NMT) proposed by \citep{zhou2019synchronous}, which incorporate right-to-left translation information directly at decoding phrase. However, the SB-NMT model is highly complex, and the decoding algorithm needs to be modified to support the bidirectional inference. Our work is mainly focused on making the decoder be able to plan for the future. Different from the above work, Our proposed method not only has a much simpler structure, but also can leave the decoding algorithm unchanged during inference.

\subsection{Knowledge Distillation}
Distillation learning is a novel transfer learning method that allows the weak model (student) to learn existing knowledge from a strong model (teacher) \citep{hinton2015distilling}. Some researchers can greatly reduce the parameters of the model without reducing the performance of the model \citep{kim2016sequence}. \citep{chen2017teacher} can train translation models with zero resources by a Teacher-Student framework. \citep{freitag2017ensemble} transfered the knowledge of the ensemble model to a single NMT model.

The most relevant works in this respect are future-aware knowledge distillation framework (FKD) \citep{zhang2019future}, Twin network \citep{serdyuk2018twin} and \citep{chen2020distilling}. FKD learns to distill future knowledge from a backward neural language model(teacher) to future-aware vectors (student) during the training phase. Twin network regularizes generative RNNs by training a backward recurrent network.  What is more, \citep{chen2020distilling} encouraged auto-regressive Seq2Seq models to plan ahead by distilling knowledge learned in BERT\citep{devlin-etal-2019-bert}, but this method requires pre-trained model BERT, which increases the complexity of the model. Different from the above model work, we use the Self-Knowledge Distillation method to learn future information from the backward decoder’s output logits, hidden layer states, so as to improve the ability of the model to plan ahead.




\section{Approach}
In this section, we introduce SBD-NMT. We first review the sequence-to-sequence(Seq2Seq) learning process in section 3.1, and then introduce the method we proposed in section 3.2, 3.3, the last section introduces Teacher Annealing mechanism.

\subsection{Sequence-to-Sequence Learning}
The Seq2Seq model \citep{sutskever2014sequence} architecture is mainly composed of an encoding structure of source sequence and a decoding structure of target sequence. The encoder of Seq2Seq encodes the input sequence into a fixed-length vector, and then the decoder generates the target sequence. 

It is usually trained via teacher forcing at each time step, and maximizes the likelihood of the next word conditioned on its previous ground-truth words, or equivalently, minimizing the cross-entropy loss as shown in Eq.\eqref{equ:seq2seq}, where $\theta$ represents the model parameters,  $\boldsymbol{X}$ represents a discrete input sequence of length \textit{M}, $\boldsymbol{Y}$ represents a discrete output sequence  of length \textit{N}, $logP_{\theta}(y_t|y_{1:t-1},\boldsymbol{X})$ represents the conditional probability when decoding the $t_{th}$ position.

\begin{equation} \label{equ:seq2seq}
\begin{split}
\mathcal{L}(\theta) & = -logP_{\theta }(\boldsymbol{Y}|\boldsymbol{X})  \\
& = -\sum_{t=1}^{N}logP_{\theta }(y_{t}|y_{1:t-1}, \boldsymbol{X}),
\end{split}
\end{equation}

Conditional probability can be learned by models such as recurrent neural network \citep{bahdanau2015neural} and Transformer \citep{vaswani2017attention}. The Transformer is based on the self-attention mechanism that is calculated in parallel, which models the global dependency of input sequence. Recently, Transformer models have achieved state-of-the-art performance on many NLP tasks including NMT, so our encoder and decoder adopt the substructure of the Transformer.

\subsection{Bidirectional Decoding}
In the decoding structure of our model, we use a forward decoder and a backward decoder.

\begin{figure}[h]
\centering
\includegraphics[width = .4\textwidth]{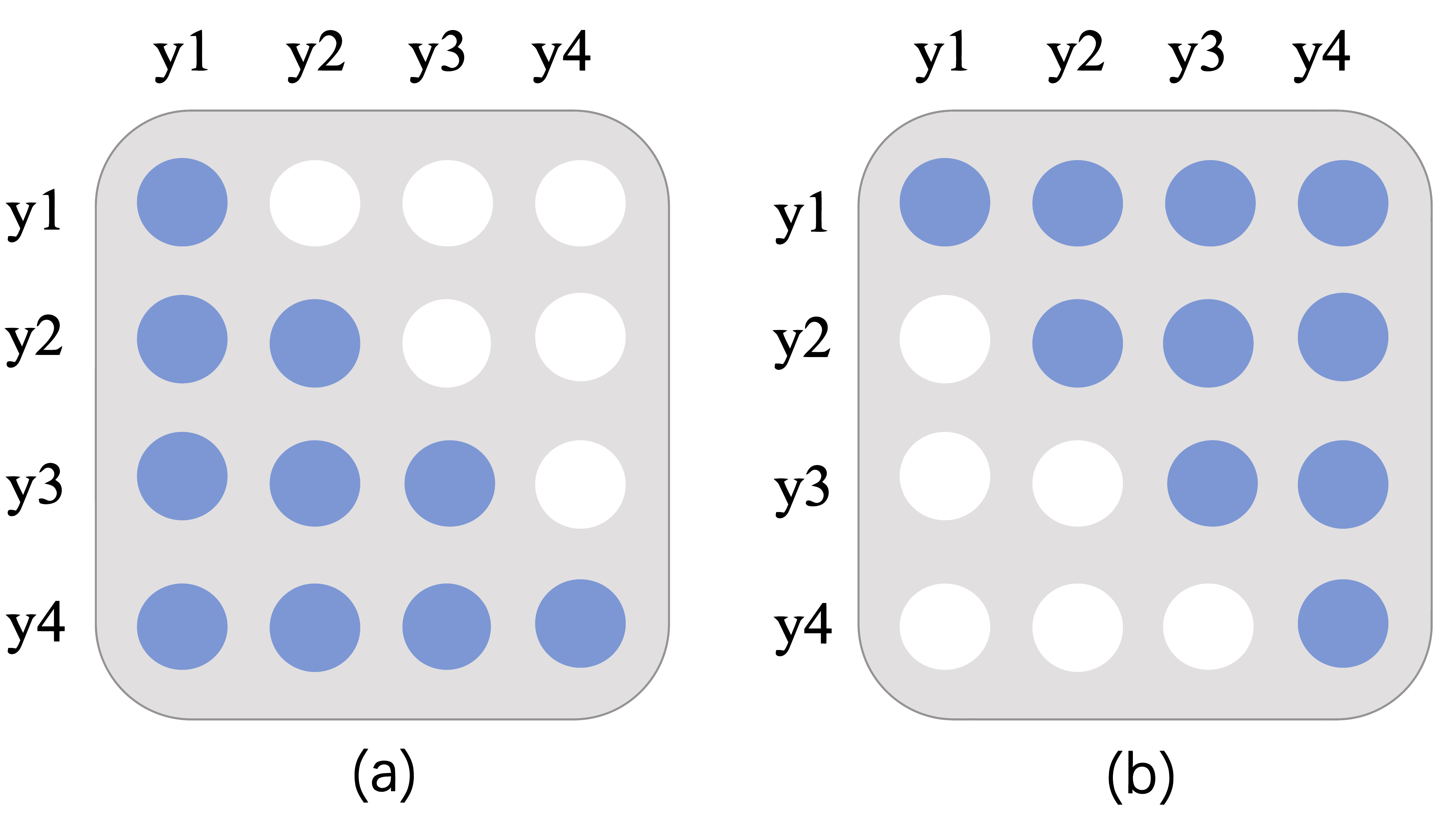}
\caption{(a):Attention mask matrix of L2R; (b):Attention mask matrix of R2L. Blank means not being attended.}
\label{fig:Figure1}
\end{figure}

The forward decoder uses the upper triangular mask matrix seeing the information on the left of $y_t$, as shown in Figure \ref {fig:Figure1}(a). We call it L2R decoder. The backward decoder models the normal language sequence from right to left, which we call R2L decoder. In the R2L decoder, we use the lower triangular mask matrix as shown in Figure \ref {fig:Figure1}(b). The two decoders are trained to generate the next symbol given the input $\boldsymbol{X}$ and the target $\boldsymbol{Y}$ from $1$ to $t-1$ or $t+1$ to $T$:

\begin{equation}
    P(y|\boldsymbol{X};\overrightarrow{\theta }) = \prod_{t=1}^{T}P(y_{t}|y_{1:t-1},\boldsymbol{X};\overrightarrow{\theta }),
\end{equation}

\begin{equation}
P(y|\boldsymbol{X};\overleftarrow{\theta }) = \prod_{t=1}^{T}P(y_{t}|y_{t+1:T},\boldsymbol{X};\overleftarrow{\theta }),
\end{equation}

L2R decoder learns information of output sequence from left to right, while R2L decoder learns the information of output sequence from right to left, which is the future information relative to L2R.
In addition, the L2R and the R2L output the probability distribution of words in every position, which has complementary information. so, we force probability distribution of $P_{L2R}$ and $P_{R2L}$ to match each other to obtain the future information.
As shown in Eq.\eqref{eq:4}, where $t$ represents the $t_{th}$ position of the output sequence, and $w$ represents the token in the vocabulary.
\begin{equation} \label{eq:4}
    \begin{split}
    & P_{R2L}(y_t = w|y_{t+1:N}, \boldsymbol{X}) 
    \\ & \qquad  \qquad \sim \\
    & P_{L2R}(y_t = w|y_{1:t-1}, \boldsymbol{X}),
    \end{split}
\end{equation}

when the two decoders are independently optimized via Maximum Likelihood Estimation (MLE), L2R decoder can not learn future information to ensure global coherence. So we use the method of Knowledge Distillation to transfer the information learned from the R2L decoder to the L2R decoder. In particular, the R2L decoder plays no role during inference, so decoding speed will not be slowed down, compared with conventional L2R NMT model.

\subsection{Self-Knowledge Distillation}

\begin{figure}[t]
\centering
\includegraphics[width = 0.5\textwidth]{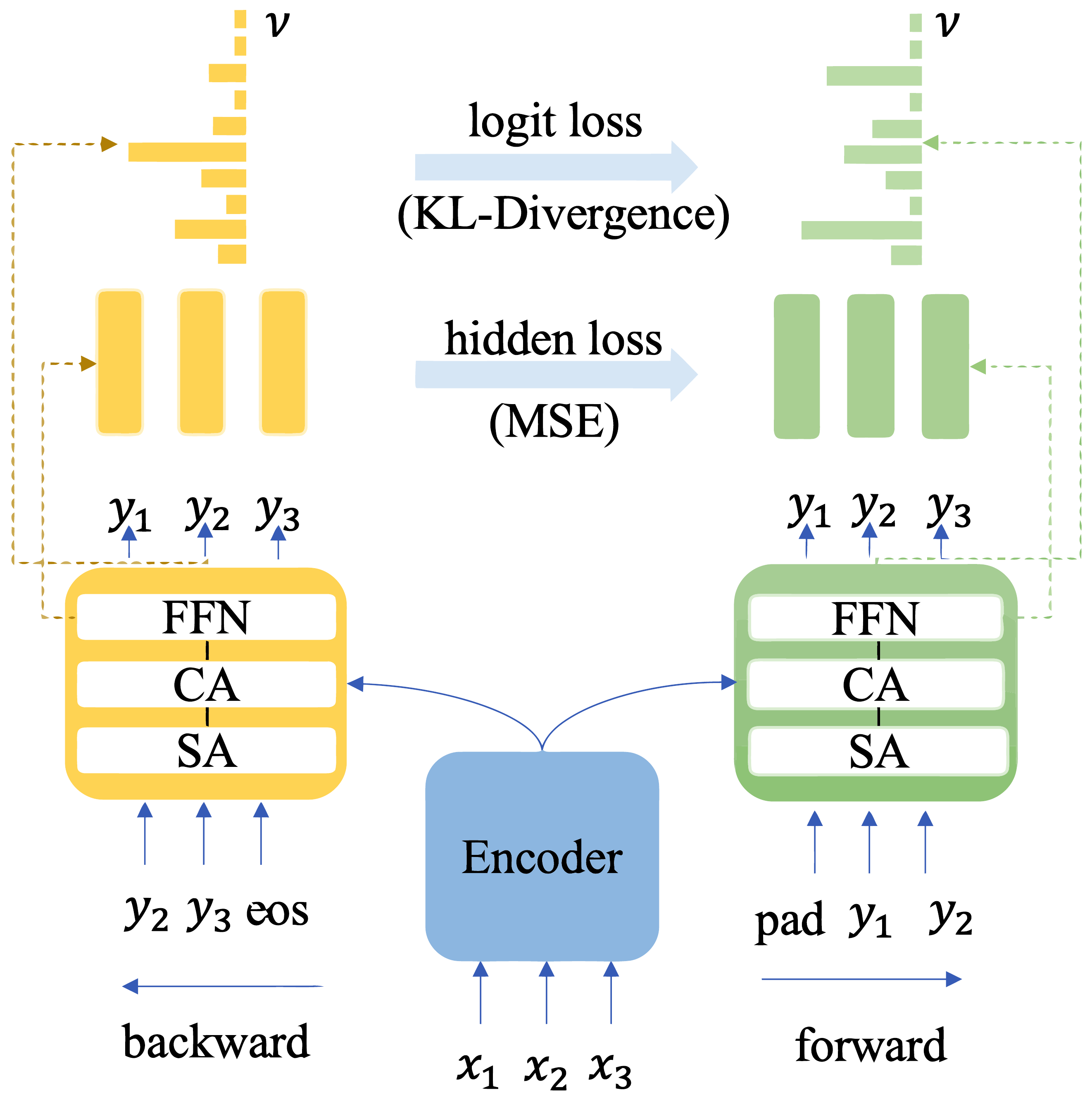}
\caption{Architecture of the proposed model, where SA is self-attention, CA is cross-attention, FFN is feedforward neural network.}
\label{fig:Figure2}
\end{figure}

This module injects the knowledge of the backward decoder into the forward decoder through knowledge distillation.  Our model uses R2L decoder as a teacher, which contains future information relative to the L2R decoder. Different from previous works that transfer teacher's knowledge layer-to-layer, we only use the logits and hidden states of the teacher's last layer, which allows more flexibility. In the process of training, the student model and the teacher model learn at the same time, so we call it Self-Knowledge Distillation. The proposed Self-knowledge Distillation includes the logit based distillation
and hidden state based distillation, which are shown in Figure \ref{fig:Figure2}. 

\subsubsection{Logit Distillation}
Logit is the prediction vector, whose dimension is the size of vocabulary and generated by the last layer of the model decoder.  It determines which symbol can be generated in the current time step. As shown in Eq.\eqref{eq:5} \eqref{eq:kl}, KL divergence is calculated from the logit probability distribution obtained by the forward decoder and backward decoder at the same position, where $T$ is the target length and $\mathcal{V}$ denotes the output vocabulary.
\begin{equation} \label{eq:5}
    \begin{split}
          \mathcal{L}_{logit} & = \\
          & \sum_{t=1}^{T}KL(P(y_{t}|y_{t+1:T},\boldsymbol{X};\overleftarrow{\theta}) || \\
           & \quad \quad \quad \quad P(y_{t}|y_{1:t-1},\boldsymbol{X};\overrightarrow{\theta})),
    \end{split}
\end{equation}

\begin{equation} \label{eq:kl}
    \begin{split}
        & KL(P(y_{t}|y_{t+1:T},\boldsymbol{X};\overleftarrow{\theta}) || P(y_{t}|y_{1:t-1},\boldsymbol{X};\overrightarrow{\theta})) \\
        & = \sum_{w \in \mathcal{V}} P(y_{t} = w |y_{t+1:T},\boldsymbol{X};\overleftarrow{\theta}) \\
        & \quad \quad \quad \times log\frac{P(y_{t}= w|y_{t+1:T},\boldsymbol{X};\overleftarrow{\theta})} {P(y_{t} = w|y_{1:t-1},\boldsymbol{X};\overrightarrow{\theta})},
    \end{split}
\end{equation}


\subsubsection{Hidden State Distillation}
In this part, we introduce the hidden state distillation method, the objective is as follows:
\begin{equation}
    \mathcal{L}_{hidden-state} = MSE(\overrightarrow{\boldsymbol{H}}\boldsymbol{W}_{h}, \overleftarrow{\boldsymbol{H}})
\end{equation}
where the matrices $\overrightarrow{\boldsymbol{H}}$ $\in$ $R^{l \times d^{'}}$ and $\overleftarrow{\boldsymbol{H}}$ $\in$ $R^{l\times d}$ refer to the hidden states of L2R and R2L network respectively. $MSE$ means the mean squared error loss function. The scalar values $d^{'}$ and $d$ indicate the hidden dimension of L2R and R2L models, and $d{'} = d$ in our model. The matrix $\boldsymbol{W}_{h}$ $\in$ $R^{d^{'} \times d}$ is a learnable linear transformation, which transforms the hidden states of L2R into the same space of the R2L hidden’s states.

Different from \citep{zhang2018asynchronous} distilling the sentence-level probability distribution, we apply these two knowledge distillation methods to force the forward decoder to learn the future information. Finally, our optimization object is shown in Eq.\eqref{eq:6}, and we construct a joint training framework.

\begin{equation} \label{eq:6}
    \begin{split}
         \mathcal{L}(\theta ) = \sum & -logP(\overleftarrow{y}|\boldsymbol{X},\overleftarrow{\theta}) \\
         & - logP(\overrightarrow{y}|\boldsymbol{X},\overrightarrow{\theta}) \\
    & + \mathcal{L}_{kd}(\overleftarrow{y},\overrightarrow{y}),
    \end{split}
\end{equation}

\begin{equation}
    \begin{split}
    \mathcal{L}_{kd} = \mathcal{L}_{logit}
    & + \mathcal{L}_{hidden-state},
    \end{split}
\end{equation}

\subsection{Teacher Annealing}

The knowledge distillation learning process is a student model imitating the teacher model. This raises the concern that the student may be limited by the teacher’s performance and not be able to substantially outperform the teacher \citep{clark2019bam}. It may lead to the student model relying heavily on the teacher model. In our SBD-NMT framework, the two distillation methods make the L2R to learn future knowledge, but in the later training stage, the model learning should pay more attention to L2R decoder training.  In order to achieve this, we propose an annealing mechanism suitable for our method. Specifically, the optimization object becomes Eq.\eqref{eq:annealing}.

\begin{equation} \label{eq:annealing}
    \begin{split}
         \mathcal{L}(\theta ) = \sum & -(1-\lambda)^{2} logP(\overrightarrow{y}|\boldsymbol{X},\overrightarrow{\theta}) \\
         & - \lambda logP(\overleftarrow{y}|\boldsymbol{X},\overleftarrow{\theta})\\
    & + (1-\lambda)\lambda \mathcal{L}_{kd}(\overleftarrow{y},\overrightarrow{y}),
    \end{split}
\end{equation}

\begin{equation} \label{eq:lambda}
   \lambda =\left\{
\begin{aligned}
&1 & where \quad c_{step}  \leq w_{step} \\
&\frac{w_{step}}{c_{step}} & where \quad c_{step} > w_{step} \\
\end{aligned}
\right.
\end{equation}

where $c_{step}$ denotes current training step, $w_{step}$ denotes warm start step. When the $c_{step}$ is less than $w_{step}$ , $\lambda$ is equal to 1, and the training objective function of the model is only $logP(\overleftarrow{y}|\boldsymbol{X},\overleftarrow{\theta})$, which can make backward decoder to learn enough knowledge. When the number of training steps is greater than $w_{step}$, $\lambda$ is equal to $\frac{w_{step}}{c_{step}}$. The effect of $logP(\overrightarrow{y}|\boldsymbol{X},\overrightarrow{\theta})$ increases, $L_{kd}(\overleftarrow{y},\overrightarrow{y})$ first increases and then decreases, whereas $logP(\overleftarrow{y}|\boldsymbol{X},\overleftarrow{\theta})$ decreases all the time.

\begin{table*}[!h]
\caption{The train/dev/test split on different MT datasets and corresponding BPE vocabularies.}
\label{fig:datasets}
\centering
\begin{tabular}{lllcccc}
\hline
\multicolumn{3}{l|}{Dataset } & \multicolumn{1}{c|}{ Train set size  } & \multicolumn{1}{c|}{ Dev set size } &  \multicolumn{1}{c|}{ Test set size } &  \multicolumn{1}{c}{Vocabulary Size} \\ \hline           
\multicolumn{3}{l|}{IWSLT 2014 De-En}  & \multicolumn{1}{c|}{160K} & \multicolumn{1}{c|}{7,283} & \multicolumn{1}{c|}{6,750} & \multicolumn{1}{c}{12.5K}  \\
\multicolumn{3}{l|}{WMT 2014 En-De}  & \multicolumn{1}{c|}{4.5M} & \multicolumn{1}{c|}{3,000} & \multicolumn{1}{c|}{3,003} & \multicolumn{1}{c}{33.7K}     \\
\multicolumn{3}{l|}{WMT 2017 En-De}  & \multicolumn{1}{c|}{5.8M} & \multicolumn{1}{c|}{2,999} & \multicolumn{1}{c|}{3,004} & \multicolumn{1}{c}{17.6K}     \\

\hline               
\end{tabular}
\end{table*}
\section{Experiments}

In this section, we conducted a set of experiments on IWSLT 2014 German-English(De-En), WMT 2014 English-German(En-De) and WMT 2017 English-German(En-De).
More experiments in appendix.

\subsection{Datasets}
WMT 2014 is the first MT dataset on which we conduct our experiment. For the En-De translation task, we use the standard WMT 2014 English-German dataset.
Specifically, we use newstest2013 as our validation set while newstest2014 as our test set. For WMT 2017 En-De translation task, we use the newstest2016 as the validation set and the newstest2017 as the test set. For IWSLT 2014 De-En, we follow the same train/dev/test split as in \citep{wu2018pay}. 


We use case-sensitive BLEU as our major evaluation metrics calculated on official tools-SacreBLEU\footnote{\url{https://github.com/mjpost/sacrebleu}}. But in order to compare with the previous results, we also use multi-bleu.perl script\footnote{\url{https://github.com/moses-smt/mosesdecoder/blob/master/scripts/generic/multi- bleu.perl}} for WMT 2014 En-De \& IWSLT 2014 De-En and case-sensitive detokenized BLEU\footnote{\url{https://github.com/moses-smt/mosesdecoder/blob/master/scripts/generic/mteval-v13a.pl}} for WMT 2017 En-De.

Based on this script\footnote{\url{https://github.com/tensorflow/models/blob/v1.9.0/official/transformer/utils/tokenizer.py}}, we apply Byte Pair Encoding (BPE)\citep{sennrich2016neural} to build the vocabulary for every individual MT dataset. The train/dev/test split on different MT datasets and corresponding BPE vocabularies are shown in Tabel \ref{fig:datasets}.

\subsection{Experimental Details}
Our implementation is based on the tensor2tensor toolkit for training and evaluating. We use 6 layer encoder and decoder, where $d_{model}=512/1024$, $P_{dropout} = 0.1/0.3$, $n_{warmup} = 4000/8000$, and 8/16 attention heads and 2048/4096 hidden feed-forward layer for base/big settings. 

We use the same warmup and decay strategy for learning rate and label smoothing as \citep{vaswani2017attention}. The proposed model is trained on one docker-machine with 8 NVIDIA P100 GPUs. Meanwhile, the batch size that specifies the approximate number of tokens (subwords) in one batch is 8,192 for every GPU. The maximum number of tokens per example is 256. In our experiments, we set $w_{step} = 30k$ for WMT 2014/2017 En-De and $w_{step} = 1k$  for IWSLT 2014 De-En. $w_{step}$ was chosen after experimentation on the development set.\footnote{The details of choosing $w_{step}$ can refer to Appendix.}.  Training took about 3 days for WMT 2014/2017 En-De and 0.5 days for IWSLT 2014 De-En.

At inference time, we use beam search with beam size of 4 and length penalty \citep{wu2016google} of 0.6 across all the models. All the hyper-parameters are tuned on the development set. For the base model, we used a single model obtained by averaging the last 5 checkpoints, which were saved every 10 minutes \citep{vaswani2017attention}. For the big model, we averaged the last 20 checkpoints. 

\subsection{Baselines}
We use the following methods for comparison.
\begin{itemize}
\item Transformer\citep{vaswani2017attention}: it is based solely on attention mechanisms, dispensing with recurrence and convolutions entirely. 
\item SB-NMT\citep{zhou2019synchronous}: it predicts its outputs using left-to-right and right-to-left decoding simultaneously and interactively. 
\item ABD-NMT\citep{zhang2018asynchronous}: it is an asynchronous bidirectional decoding for NMT, which equipped the conventional attentional encoder-decoder NMT model with a backward decoder.
\item Rerank-NMT\citep{liu2016agreement}: it first runs beam search for left-to-right and right-to-left NMT models independently to obtain two k-best lists, and then re-score the union of two k-best lists using the joint model to find the best candidate.
\item TwinNMT\citep{zhang2019improving}: it trains a “backward” recurrent network to generate a given sequence in reverse order, and encourages states of the forward model to predict cotemporal states of the backward model.
\item FKD\citep{zhang2019future}: it distills future knowledge from a backward neural language model (teacher) to future-aware vectors (student) during the training phase.
\item RT\citep{zhang2019your}: it introduce two Kullback-Leibler divergence regularization terms into the NMT training objective to reduce the mismatch between output probabilities of L2R and R2L models.
\item \citep{zheng2018modeling}: it separates the source information into two parts: translated PAST contents and untranslated FUTURE contents what fed to both the attention model and the decoder states.
\item \citep{chen2020distilling}: it  distills knowledge learned in BERT can encourage
auto-regressive Seq2Seq models to plan ahead.
\end{itemize}

\subsection{Experiment Results}

In this section, we present the results of our model on different datasets. Meanwhile, the statistics are displayed as tables and charts, which show that our model signiﬁcantly improves the strong Transformer baseline across all datasets.


\begin{table}[!h]
\caption{The experiment results of WMT 2014 En-De translation, measured by multi-bleu. The method is marked token \textsuperscript{*} from \citep{zhou2019synchronous}.}
\label{tab1}
\centering 
\begin{tabular}{lllc}
\hline
\multicolumn{3}{l|}{WMT 2014 En-De}   & \multicolumn{1}{c}{TEST} \\ \hline
\multicolumn{4}{c}{Our Implementations}                                 \\ \hline
\multicolumn{3}{l|}{Transformer (base)}  & 27.66                            \\
\multicolumn{3}{l|}{Our Model (base)}  &28.36                       \\ \hline
\multicolumn{3}{l|}{Transformer (big)}   & 28.88                 \\ 
\multicolumn{3}{l|}{Our Model (big)}   & \textbf{29.51}                     \\  \hline
\multicolumn{4}{c}{Other Reported Results}                                \\ \hline

\multicolumn{3}{l|}{SB-NMT\textsuperscript{*}}             & 29.21  \\
\multicolumn{3}{l|}{ABD-NMT\textsuperscript{*}}            & 28.22      \\
\multicolumn{3}{l|}{Rerank-NMT\textsuperscript{*}}         & 27.81 \\ 

\hline 
\end{tabular}
\end{table}
\subsubsection{Results on English-German Translation}
For large-scale WMT 2014, the experiments result where the En-De translation tasks conduct on is shown in Table \ref{tab1}. Specifically, the proposed model(big) significantly outperforms Rerank-NMT, ABD-NMT, SB-NMT by 1.7, 1.29, and 0.3 BLEU points, respectively. Compared with Rerank-NMT in which two decoders are relatively independent, ABD-NMT which is a two-stage decoder architecture, which gets final translation based on source sentence and previous generated R2L translation, and SB-NMT where sharing decoder parameters between two directional decoders may add constraints to decoding in two directions, our proposed model achieves substantial improvements over them, which indicates that the Self-Knowledge Distillation between bidirectional decoder behaves better in acting as an effective regularization method by leveraging  backward decoder's looking into the future ability.

\begin{table}[!h]
\caption{The experiment results of WMT 2017 En-De translation, measured by case-sensitive detokenized BLEU and SacreBLEU. The score in quota is the SacreBLEU. \textsuperscript{$\dagger$} from \citep{zhang2019future}. \textsuperscript{$\ddagger$} means that the result is taken from the corresponding paper. \textsuperscript{*} from \citep{zhang2019regularizing}. BT denotes back-translation
method.}
\label{tab_wmt17}
\centering
\begin{tabular}{lllc}
\hline
\multicolumn{3}{l|}{WMT 2017 En-De } & \multicolumn{1}{c}{TEST}  \\ \hline
\multicolumn{4}{c}{Our Implementations}    \\ \hline
\multicolumn{3}{l|}{Transformer (base)}  & \multicolumn{1}{c}{28.71(26.70)} \\
\multicolumn{3}{l|}{Our Model (base)}    & \multicolumn{1}{c}{29.21(27.56)}   \\ \hline
\multicolumn{3}{l|}{Transformer (big)}  & \multicolumn{1}{c}{29.47(27.49)} \\
\multicolumn{3}{l|}{Our Model (big)}    & \multicolumn{1}{c}{30.7(28.53)}   \\ \hline
\multicolumn{4}{c}{Other Reported Results}   \\ \hline
\multicolumn{3}{l|}{Transformer(base)+FKD\textsuperscript{$\dagger$}}  & \multicolumn{1}{c}{28.80}    \\ 
\multicolumn{3}{l|}{Transformer(big)+BT+RT\textsuperscript{*}}  & \multicolumn{1}{c}{(\textbf{29.46})} \\
\multicolumn{3}{l|}{Transformer(base)+RT\textsuperscript{*}}  & \multicolumn{1}{c}{(27.18)} \\

\multicolumn{3}{l|}{\citep{zheng2018modeling}\textsuperscript{$\ddagger$}} & \multicolumn{1}{c}{24.3}   \\
\multicolumn{3}{l|}{TwinNMT\textsuperscript{$\dagger$}}  & \multicolumn{1}{c}{23.9}    \\

\hline        
\end{tabular}
\end{table}

For WMT 2017 En-De, as Table \ref{tab_wmt17} shows, our model(base) achieve the highest performance on SacreBLEU and detokenized BLEU, compared with other reported results except Transformer+BT+RT. Different from our model, Transformer+RT uses translation candidates generated by bi-directional independent models to regularize each other. Transformer+BT+RT uses back-translations method\citep{sennrich2015improving} which improve neural machine translation models with monolingual data. Transformer+FKD distills future knowledge from a backward neural language model trained only by target data. The results show \citep{zheng2018modeling} and TwinNMT are not as good as ours in capturing subtle long-term dependencies and improving translation quality.

\subsubsection{Results on German-English Translation}
For IWSLT 2014 De-En, as Table \ref{tab_iwslt2014deen} shows, our model(big) achieves 35.92 BLEU points. Although \citep{chen2020distilling} ensure global coherence by the knowledge distillation, it requires additional large-scale pre-trained model and fine-tuning on MT datasets which is complex and multi-stage training. Instead, the proposed model plans for future token prediction by self-knowledge distillation and jointly training on MT datasets.



\begin{table}[!h]
\caption{The experiment results of IWSLT 2014 De-En translation tasks on the test set, measured by multi-bleu. \textsuperscript{$\ddagger$} means that the result is taken from the corresponding paper.}

\label{tab_iwslt2014deen}
\centering
\begin{tabular}{lllc}
\hline
\multicolumn{3}{l|}{IWSLT 2014 De-En} & \multicolumn{1}{c}{TEST} \\ \hline
\multicolumn{4}{c}{Our Implementations}                                                                                  \\\hline
\multicolumn{3}{l|}{Transformer (base)}          & 34.35                            \\
\multicolumn{3}{l|}{Our Model (base) }    &       35.36                        \\ \hline
\multicolumn{3}{l|}{Transformer (big)}    & 34.92                  \\ 
\multicolumn{3}{l|}{Our Model (big)}       & \textbf{35.92}                    \\ \hline
\multicolumn{4}{c}{Other Reported Result}                                                                               \\ \hline
\multicolumn{3}{l|}{\citep{chen2020distilling}\textsuperscript{$\ddagger$}}               &  35.63                            
\\\hline               
\end{tabular}
\end{table}




\begin{table}[!h]
\caption{The ablation test is conducted on IWSLT 2014 De-En translation tasks, measured by SacreBLEU. $Trm(\rightarrow)$ means Transformer(base) with forward decoder. $Trm(\leftarrow)$ means Transformer(base) with backward decoder.$Trm(\rightarrow \& \leftarrow)$  means Transformer(base) with forward and backward decoders.
}
\label{ab:tab1}
\centering
\begin{tabular}{lllcc}
\hline
\multicolumn{3}{l|}{IWSLT 2014 De-En } & \multicolumn{1}{c|}{VALID}& \multicolumn{1}{c}{TEST} 
\\ \hline           
\multicolumn{3}{l|}{$Trm(\rightarrow)$}  & \multicolumn{1}{c|}{38.2} & \multicolumn{1}{c}{37.2} 
\\
  
\multicolumn{3}{l|}{$Trm(\leftarrow)$}   & \multicolumn{1}{c|}{38.0} & \multicolumn{1}{c}{37.0}                    
\\

\multicolumn{3}{l|}{$Trm(\rightarrow \& \leftarrow)$}   & \multicolumn{1}{c|}{38.4} & \multicolumn{1}{c}{37.5}                    
\\

\multicolumn{3}{l|}{Our Model}            & \multicolumn{1}{c|}{39.2} & \multicolumn{1}{c}{38.2}                    
\\ \hline
\multicolumn{3}{l|}{ - Teacher Annealing}            & \multicolumn{1}{c|}{38.8} & \multicolumn{1}{c}{37.8} 
\\


\multicolumn{3}{l|}{ - hidden KD}            & \multicolumn{1}{c|}{38.9} & \multicolumn{1}{c}{37.9} 
\\

\multicolumn{3}{l|}{ - logit KD}            & \multicolumn{1}{c|}{38.6} & \multicolumn{1}{c}{37.8} 
\\ \hline


\end{tabular}
\end{table}




\begin{table*}[h!]
\caption{Qualitative examples from IWSLT 2015 Chinese-English translation task. L2R model perform worse \textcolor{red}{in the last part of sentence}, whereas R2L translates worse \textcolor{blue}{in the first part of the sentence}. }
\label{case1}
\begin{tabular}{l|l|l|l|l|l|l|l|l|l|l}
\hline
\multicolumn{2}{l|}{Source}   & \multicolumn{9}{l}{\begin{tabular}[c]{@{}l@{}}
nathaniel cóng zhū lì yà yīn yuè xué yuàn chuò xué , tā wán quán bēng kuì le\\ 30 nián hòu de tā liú luò zài luò shān jī skid row dà jiē shàng wú jiā kě guī\end{tabular}} \\ \hline
\multicolumn{2}{l|}{Golden Target} & \multicolumn{9}{l}{\begin{tabular}[c]{@{}l@{}}nathaniel dropped out of juilliard , he suffered a complete breakdown , and 
30 \\ years later he was living homeless on the streets of skid row in
downtown los \\ angeles .\end{tabular}}                                                                                                                                                                                                           \\ \hline
\multicolumn{2}{l|}{L2R Model}     & \multicolumn{9}{l}{\begin{tabular}[c]{@{}l@{}}nathaniel dropped out of juilliard school , and he absolutely collapsed 30 years \\later , and \textcolor{red}{ he landed in l.a. skid row on the streets .}\end{tabular}}                                                                                                                                                                                                                       \\ \hline
\multicolumn{2}{l|}{R2L Model}     & \multicolumn{9}{l}{\begin{tabular}[c]{@{}l@{}}\textcolor{blue}{nathaniel pulled out as a school dropout from the school of juilliard music ,} and \\ he completely broke down , and 30 years later he was in l.a. skid row who was \\ homeless on the street .\end{tabular}}                                                                                                                                                                                                                          \\ \hline
\multicolumn{2}{l|}{Our Model}     & \multicolumn{9}{l}{\begin{tabular}[c]{@{}l@{}}nathaniel drowned out of juilliard music school , and he collapsed , and 30 years \\
later he landed on skid row in los angeles , homeless on the street .\end{tabular}}                                                                                                                                                                         \\ \hline
\end{tabular}
\end{table*}


\subsection{Ablation Study}

The Knowledge Distillation (KD) module and Teacher Annealing on NMT are our key contributions in this work. To investigate their effectiveness, we conduct an ablation test on IWSLT 2014 De-En translation task. 

As Table \ref{ab:tab1} shows, our model achieves 0.7/1.0 BLEU (VALID/TEST) improvement over $Trm(\rightarrow)$.   $Trm(\leftarrow)$ obtains 0.2/0.2 BLEU (VALID/TEST) lower than $Trm(\rightarrow)$, which is similar to the experimental phenomenon\citep{zhou2019synchronous}. $Trm(\rightarrow \& \leftarrow)$ simultaneously learns forward and reverse translation tasks, it achieves 0.2/0.3 BLEU (VALID/TEST) improvement over $Trm(\rightarrow)$. When our model is separately removed from the Hidden KD and the logit KD, the performance drops separately by 0.3/0.4 BLEU (VALID/TEST) and 0.6/0.4 BLEU (VALID/TEST). 
When removing the Teacher Annealing, the performance drops by 0.4/0.4 BLEU(VALID/TEST). The result shows that the KD module and Teacher Annealing are both helpful for performance improvement.


\subsection{Effect of Different Length Sentences}




\begin{figure}[h]
\centering
\includegraphics[width = .5\textwidth]{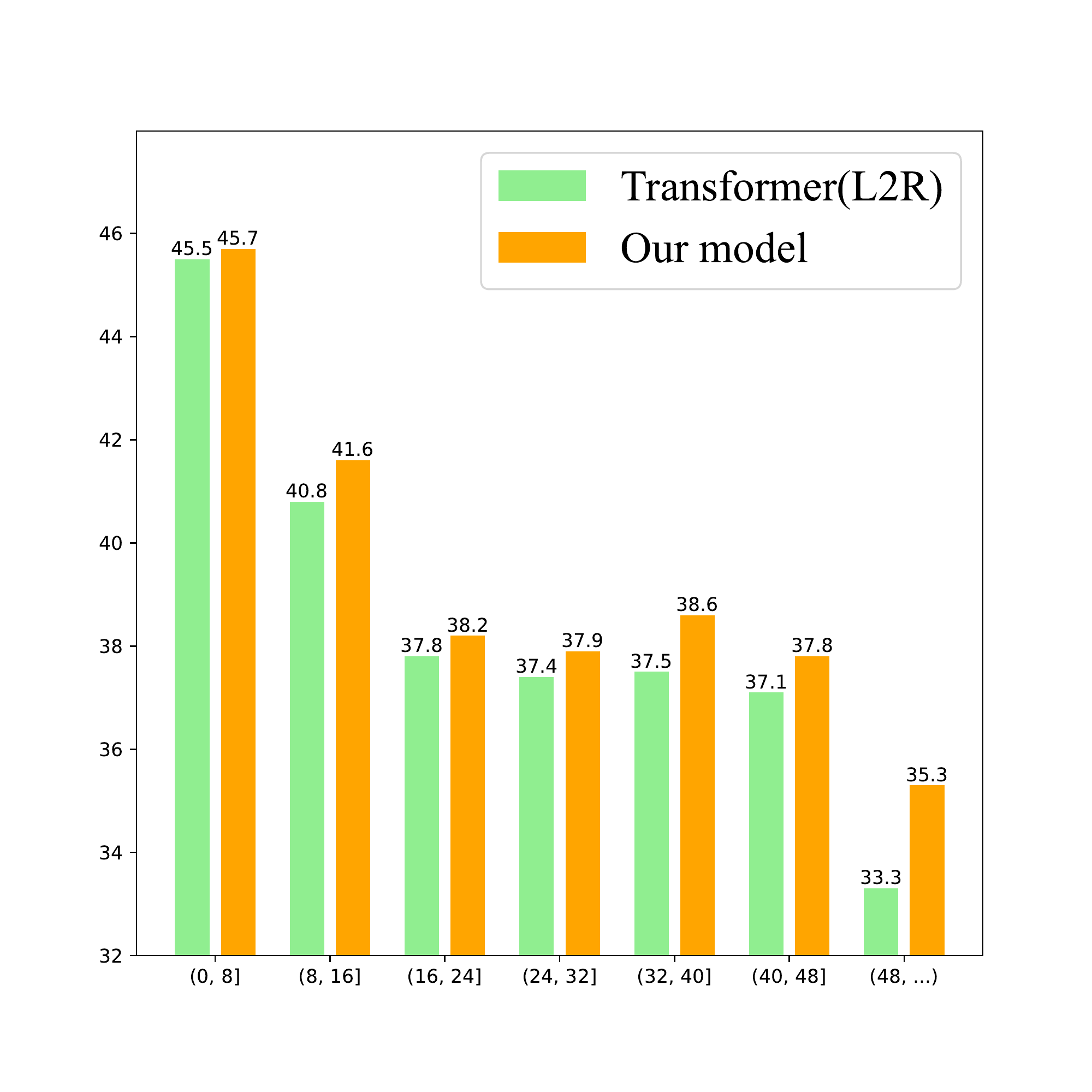}
\caption{SacreBLEU scores on IWSLT German-English for
different source lengths.}
\label{fig:Figure3}
\end{figure}

As shown in Figure \ref{fig:Figure3}. We group source sentences of similar number of words and compute a BLEU score per group. We can see a shared trend that the proposed model gains higher BLEU points on longer sentence over Transformer(L2R). In fact, self-knowledge distillation  with bidirectional decoder boosts translation performance on all source sentence groups. It further shows that our method is effective by taking advantage of a backward decoder looking into the future, which can act as an effective regularization method.

\subsection{Case Study}
Table \ref{case1} gives a examples to show translations from different models. When the source sentence is long, translation model tends to suffer from long-range dependency issue. Specifically, the L2R translation model fails to translate the last source sentence, and the R2L translation model mistakenly combines the first and second subsentence. In contrast, our model succeeds in overcoming all these problems
We attribute this to the introduction of by Self-Knowledge Distillation which distilling future knowledge from the backward decoder looking into the future ability to the forward decoder. Self-Knowledge Distillation can act as an effective regularization method, capturing subtle long-term dependencies that ensure global coherence and in consequence boost model performance. Other cases are presented in Appendix.




\section{Conclusion and Future Work}

In this work, we proposed a novel NMT architecture, which is called Self-Knowledge Distillation with Bidirectional Decoder for NMT. Our model can fully focuse on global coherence rather than local structure by implicitly forcing the forward decoder to hold information about the longer-term future contained in the backward decoder.
Experiments on WMT 2014/2017 En-De translation task and IWSLT 2014 De-En translation tasks demonstrate the effectiveness of our model
in improving both translation quality.
In the future, we would like to experiment with more language
pairs and apply our method to other sequence-to-sequence tasks, such as abstractive summarization, and
image captioning. 


\bibliographystyle{acl_natbib}
\bibliography{SBD-NMT}
\clearpage






\appendix

\section{Training}
\label{sec:appendix}

\begin{algorithm}
        \caption{Algorithm for Training Our Model}
        \KwIn{D=$(\boldsymbol{X}^n,\boldsymbol{Y}^n)_{n=1}^N$ ;Trainable parameter set: $\theta_{L2R},\theta_{R2L}, \theta_{KL}$} 
        \KwOut{Trained full NMT model, $\theta_{L2R}$ }
        \While{Not converged and Not reaching the maximum training epoch}{
             \For{each mini-batch(x,y) $\epsilon$ D} 
             {
             	   Sample sentence pairs $x^{n}$,$y^{n}$ from  D; \\
	   	   Generate translation $y_1,y_2,\ldots, y_{T}$ by translation model $P(y_{t}|y_{t+1:T},\boldsymbol{X};\overleftarrow{\theta })$ and $P(y_{t}|y_{1:t-1},\boldsymbol{X};\overrightarrow{\theta })$;\\
		   Calculate the loss value by Eqution 7; \\
		   Update model parameter: \\
                   	 $\quad \quad \quad \theta_{R2L}$ $\leftarrow$ opt($\theta_{R2L}$,  $\frac{\partial L(\theta) }{\partial \theta_{R2L}}$); \\
                   	 $\quad \quad \quad \theta_{L2R}$ $\leftarrow$ opt($\theta_{L2R}$, $\frac{\partial L(\theta) }{\partial \theta_{L2R}}$ ); \\
	 		$\quad \quad \quad \theta_{KL}$ $\leftarrow$ opt($\theta_{KL}$,  $\frac{\partial L(\theta) }{\partial \theta_{KL}}$); \\
                            }
              }
\end{algorithm}
We show the overall training procedure in Algorithm 1, where we construct parallel corpus pairs as a training set, and learn model parameters through translation tasks and distillation tasks. In addition, we distinguish parameters $\theta_{R2L}$ in the backward decoder and $\theta_{kd}$ in the knowledge distillation module from other parameters $\theta_{L2R}$ in the conventional encoder-decoder NMT model. Specifically, we first optimize the R2L module for ensuring the backward decoder to learn enough knowledge. Then we optimize the $\theta_{kd}$ $\theta_{L2R}$ and transfer the knowledge learned from $\theta_{R2L}$ to $\theta_{L2R}$. Our algorithm implementation process is very simple, the generation task and the distillation tasks are constructed as a joint learning framework, and the parameters are learned in each mini-batch.

\section{Details of Choosing $w_{step}$}

\begin{figure}[h]
\centering
\includegraphics[width = .5\textwidth]{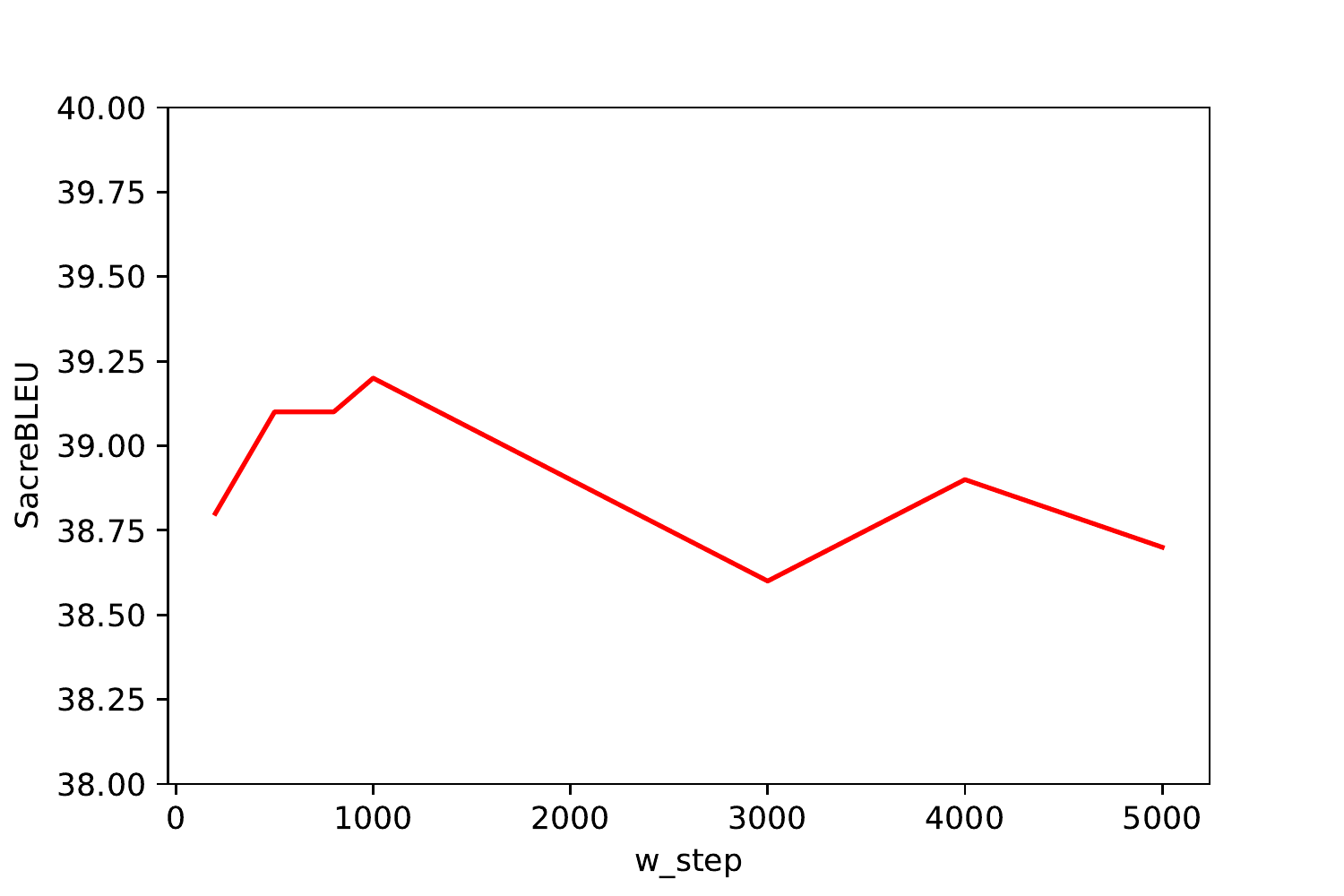}
\caption{Performance changes of different $w_{step}$}
\label{fig:Figure4}
\end{figure}

We conducted parameter optimization experiments for $w_{step}$ on the IWSLT 2014 De-En validation set. As shown in the Figure \ref{fig:Figure4}, as $w_{step}$ increases, the performance is gradually improved, and is best in 1K. When $w_{step}$ more than 1K, the performance presents a descendent trend. So we set $w_{step}=1k$ in the IWSLT De-En experiment. For the WMT 2014/2017 En-De datasets, since they are about 30 times the size of IWSLT 2014 De-En, we set $w_{step}=30k$.

\section{More experiments}

\begin{table*}[!h]
\caption{The train/dev/test split on different MT datasets and corresponding BPE vocabularies.}
\label{fig:datasets1}
\centering
\begin{tabular}{lllcccc}
\hline
\multicolumn{3}{l|}{Dataset } & \multicolumn{1}{c|}{ Train set size  } & \multicolumn{1}{c|}{ Dev set size } &  \multicolumn{1}{c|}{ Test set size } &  \multicolumn{1}{c}{Vocabulary Size} \\ \hline           

\multicolumn{3}{l|}{IWSLT 2015 En-Vi/Vi-En}  & \multicolumn{1}{c|}{133K} & \multicolumn{1}{c|}{1,553} & \multicolumn{1}{c|}{1,268} & \multicolumn{1}{c}{9.3K}  \\
\multicolumn{3}{l|}{IWSLT 2015 Zh-En}  & \multicolumn{1}{c|}{210K} & \multicolumn{1}{c|}{887} & \multicolumn{1}{c|}{5,437} & \multicolumn{1}{c}{11.5K}  \\
\hline               
\end{tabular}
\end{table*}

\begin{figure*}[!h]

\includegraphics[width=16.cm]{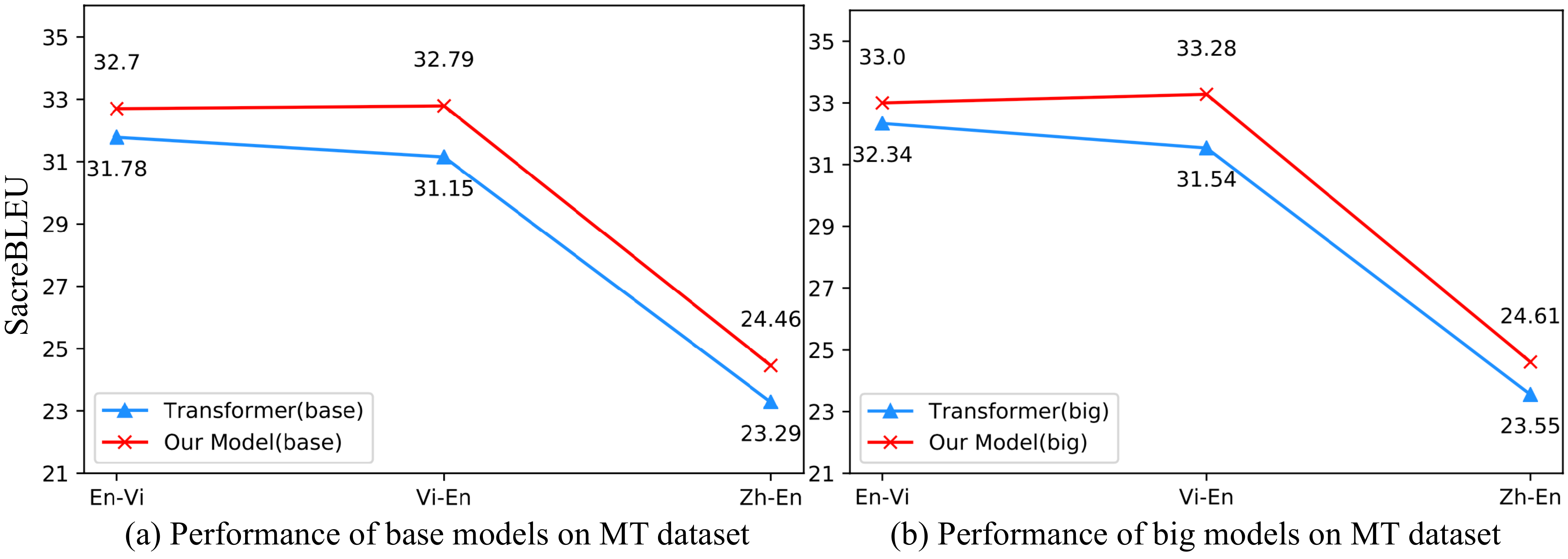}
\caption{Performance comparison of Our Model and Transformer on IWSLT 2015 En-Vi, Vi-En and Zh-En MT datasets.}
\label{exp:sacrebleu}
\end{figure*}

In this section, we conducted a set of experiments on Chinese-English (Zh-En), Vietnamese-English(Vi-En), and English-Vietnamese(En-Vi) 
translation tasks.
For IWSLT 2015\footnote{\url{http://workshop2015.iwslt.org/}}, we apply three translation tasks, Vi-En, En-Vi, and Zh-En. For En-Vi and Vi-En, the training/validation/test sets, 
follow \citep{chen2020distilling}. Zh-En corpus consists of Simplified Chinese and English pairs. The train/dev/test split on different MT datasets and corresponding BPE vocabularies are shown in Tabel \ref{fig:datasets1}.

Figure \ref{exp:sacrebleu} shows that the proposed model signiﬁcantly outperforms strong Transformer baselines on SacreBLEU. The results above demonstrate that our Self-Knowledge Distillation module has an improvement on multiple MT tasks, since the only difference between our model and Transformer is the Self-Knowledge Distillation module.

\section{Other Case Study}

Table \ref{case2} compares the translations from different models on IWSLT 2015 Zh-En MT dataset. When the source sentence is long, translation model tends to suffer from strong local dependency and exposure bias issues. Specifically, the L2R translation model fails to translate the last source sentence, and the R2L translation model mistakenly combines the first and second sub-sentence.

\label{sec:appendix}
\begin{table*}[]

\caption{Qualitative examples from IWSLT 15 Chinese-English translation task. L2R model perform worse \textcolor{red}{in the last part of sentence}, whereas R2L translates worse \textcolor{blue}{in the first part of the sentence}. } 
\label{case2}
\begin{tabular}{l|l|l|l|l|l|l|l|l|l|l}
\hline

\multicolumn{2}{l|}{Source} & \multicolumn{9}{l}{\begin{tabular}[c]{@{}l@{}}
wǒ xiàn zài yào jiǎng de zhè gè gù shì nǐ men hěn duō rén yǐ jīng zhī dào le yīn wèi\\  steve  de zhuān lán hòu lái chéng le yī běn shū de jī chǔ , rán hòu yòu bèi pāi chéng \\diàn yǐng qí zhōng  robert downey jr  bàn yǎn steve lopez jamie foxx  bàn yǎn le nathaniel \\anthonyayer  tā yuán běn shì zhū lì yà yīn yuè xué yuàn péi xùn de shuāng chóng bèi sī \\shǒu bù liào tā de zhí yè shēng yá què yīn wèi huàn shàng piān zhí xíng jīng shén \\ fēn liè zhèng ér bù xìng zhōng duàn 

\end{tabular}}  \\ \hline

\multicolumn{2}{l|}{Golden Target} & \multicolumn{9}{l}{\begin{tabular}[c]{@{}l@{}}and i'm telling a story that many of you know , because steve's columns became \\ the basis for a book , which was turned into a movie , with robert downey jr. \\ acting as steve lopez , and jamie foxx as nathaniel anthony ayers , the \\ juilliard-trained double bassist whose promising career  was cut short by a tragic \\ affliction with paranoid schizophrenia .\end{tabular}}                                                                                                                                                                                                           \\ \hline
\multicolumn{2}{l|}{L2R Model}     & \multicolumn{9}{l}{\begin{tabular}[c]{@{}l@{}} and i'm going to tell you a lot of you already know , because steve's column has \\  become the basis of a book , and then it's made into a movie , and robert \\ donnelly plays steve jamie faye , who played nathaniel yann , who was a \\ graduate student at the juilliard school ,\textcolor{red} {and had a career in which he had l}\\ \textcolor{red}{ earned that he had a mighty scheme} .\end{tabular}}                                                                                                                                                                                                                       \\ \hline
\multicolumn{2}{l|}{R2L Model}     & \multicolumn{9}{l}{\begin{tabular}[c]{@{}l@{}}\textcolor{blue}{now many of you already know that} , because steve's column then became the \\ foundation of a book and then made it into a movie , and robert wharev. pez \\ jeremy orthodox play nathaniel anthony downey , \
 who was individually trained   \\  by the wife of juley , who had a career in schizophrenia , and unfortunately he \\  suffered from schizophrenia .\end{tabular}}                                                                                                                                                                                                                          \\ \hline
\multicolumn{2}{l|}{Our Model}   & \multicolumn{9}{l}{\begin{tabular}[c]{@{}l@{}} now , i'm going to tell you this story that many of you already know , because \\  steve's column became the foundation of a book , and then it's made into a \\  movie , and robert donney jr plays steve lozo jami fox , who plays nathaniel \\  anthony yare , who's otherwise a double bassist at the juilliard music academy , \\  and he's not prepared for his career to be distracted with schizophrenia.\end{tabular}}                                                                                                                                                                         \\ \hline
\end{tabular}
\end{table*}






\end{document}